\documentclass{article}

 \usepackage[preprint]{neurips_2025}

\usepackage{wrapfig}

\usepackage[utf8]{inputenc}
\usepackage[T1]{fontenc}
\usepackage{hyperref}
\usepackage{url}
\usepackage{booktabs}
\usepackage{amsfonts}
\usepackage{nicefrac}
\usepackage{microtype}
\usepackage[dvipsnames]{xcolor}

\usepackage{tcolorbox}

\title{Preference-Based Gradient Estimation for ML-Guided Approximate Combinatorial Optimization}

\author{%
  Arman Mielke \\
  ETAS Research \\
  Stuttgart, Germany \\
  \texttt{arman.mielke@etas.com} \\
  \And
  Uwe Bauknecht \\
  ETAS Research \\
  Stuttgart, Germany \\
  \texttt{uwe.bauknecht@etas.com} \\
  \AND
  Thilo Strauss \\
  School of AI and Advanced Computing \\
  Xi'an Jiaotong-Liverpool University \\
  China \\
  \texttt{thilo.strauss@xjtlu.edu.cn} \\
  \And
  Mathias Niepert \\
  Computer Science Department \\
  University of Stuttgart \\
  Stuttgart, Germany \\
  \texttt{mathias.niepert@ki.uni-stuttgart.de} \\
}

\usepackage{amsmath}
\usepackage{amssymb}
\usepackage{amsthm}
\usepackage{graphicx}
\usepackage{bm}
\usepackage{subcaption}
\usepackage[printonlyused]{acronym}
\usepackage{algorithm}
\usepackage{algpseudocode}

\newacro{PBGE}[PBGE]{preference-based gradient estimation}
\newacro{GNN}[GNN]{graph neural network}
\newacro{CO}[CO]{combinatorial optimization}
\newacro{TSP}[TSP]{travelling salesman problem}
\newacro{RL}[RL]{reinforcement learning}
\newacro{MLP}[MLP]{multi-layer perceptron}
\newacro{I-MLE}[I-MLE]{implicit maximum likelihood estimator}
\newacro{BCE}[BCE]{binary cross entropy}
\newacro{LLM}[LLM]{large language model}

\newtheorem{theorem}{Theorem}

\newtheorem{corollary}[theorem]{Corollary}
\theoremstyle{definition}

\def\gG{{\mathcal{G}}}

\DeclareMathOperator*{\argmin}{arg\,min}

\begin{document}

\maketitle

\begin{abstract}
\Ac{CO} problems arise across a broad spectrum of domains, including medicine, logistics, and manufacturing. While exact solutions are often computationally infeasible, many practical applications require high-quality solutions within a given time budget. To address this, we propose a learning-based approach that enhances existing non-learned approximation algorithms for \ac{CO}.
Specifically, we parameterize these approximation algorithms and train \acp{GNN} to predict parameter values that yield near-optimal solutions. Our method is trained end-to-end in a self-supervised fashion, using a novel gradient estimation scheme that treats the approximation algorithm as a black box.
This approach combines the strengths of learning and traditional algorithms: the \ac{GNN} learns from data to guide the algorithm toward better solutions, while the approximation algorithm ensures feasibility.
We validate our method on two well-known combinatorial optimization problems: the \ac{TSP} and the minimum $k$-cut problem. Our results demonstrate that the proposed approach is competitive with state-of-the-art learned \ac{CO} solvers.
\end{abstract}

\section{Introduction}

The design and analysis of approximation algorithms for \acf{CO} often focuses on improving worst-case performance.
However, such worst-case scenarios may rarely occur in real-world settings. Learning-based approaches offer the advantage of adapting to the distribution of problem instances encountered in practice.
\Acfp{GNN} are often the method of choice since most \ac{CO} problems are either defined on graphs or admit graph-based formulations. Since neural networks cannot directly predict solutions guaranteed to be feasible, generic algorithms such as Monte Carlo tree search and beam search are often used to decode their outputs. However, these methods are impractical to use during training because of their prohibitively long runtime.  Consequently, omitting them during training introduces a discrepancy between training and inference, which can severely impact the optimality gap and generalization.

To overcome these limitations, we propose to use \ac{GNN}s to augment highly efficient approximation algorithms that are commonly available for many \ac{CO} problems.
The \ac{GNN}s are trained to predict parameters for the approximation algorithms that influence their behavior.
Specifically, we predict edge weights for the input graphs and then apply the approximation algorithms to this modified graph.
By choosing fast approximation algorithms, we can incorporate them into the training loop and use the exact same pipeline during training and inference.
Our method combines the strengths of data-driven and established heuristic algorithms: the \ac{GNN}s learn to perform well on frequently encountered instances, and the approximation algorithm, in turn, ensures the feasibility of the resulting solutions.

Once trained, the GNN-augmented approximation algorithms serve as drop-in replacements, preserving efficiency and usability while improving solution quality. Indeed, we provide a theoretical guarantee that, for a broad class of approximation algorithms, a suitably trained upstream \ac{GNN} can learn parameters that render the approximation algorithm optimal.

\begin{figure*}
        \centering
        \includegraphics[width=\textwidth]{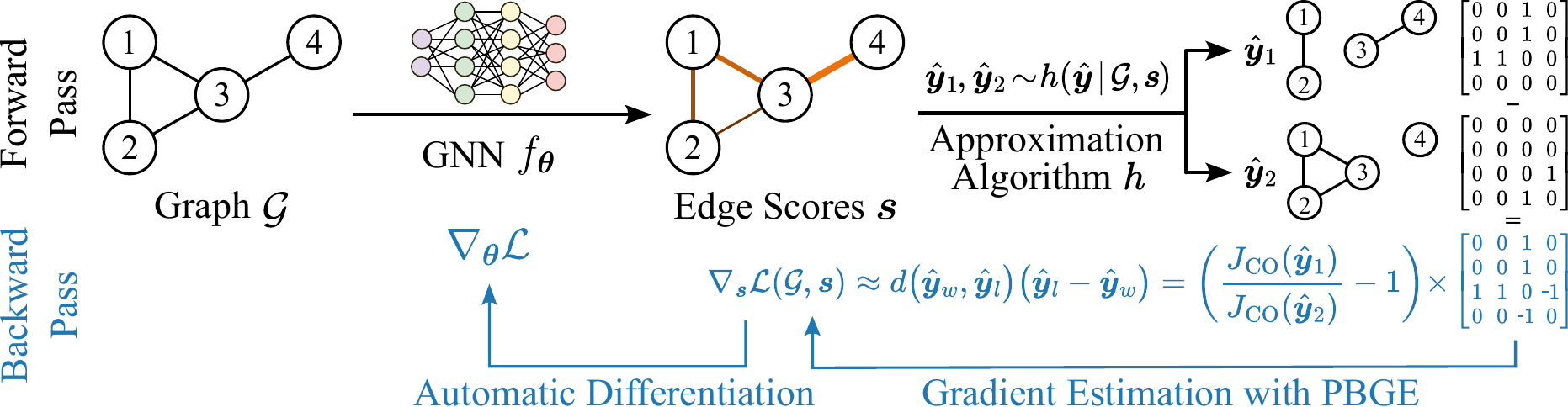}
        \caption{
            Overview of the proposed framework for ML-guided approximate combinatorial optimization.
            In the forward pass, the \ac{GNN} predicts edge scores that are used as input parameters for a \ac{CO} approximation algorithm.
            The approximation algorithm operates on these parameters by using them to scale the graph's edge weights and running an off-the-shelf approximation algorithm on the modified graph to obtain a solution $\hat{\bm{y}}$.
            Since the CO approximation algorithm is not differentiable in general, a gradient estimation scheme is used in the backward pass.
        }
        \label{fig:approach-overview}
    \end{figure*}

Since the approximation algorithms return discrete solutions, the gradients of their outputs with respect to their inputs are zero almost everywhere. 
We therefore need to apply gradient estimation techniques such as the score function and straight-through estimators \citep{reinforce, straight-through-estimator} to backpropagate through the algorithms. 
Since existing gradient estimators cannot be used in a self-supervised setting or have high variance, we propose \acf{PBGE}, which estimates gradients through a comparison of solutions sampled from the given approximation algorithm.
This approach enables fully self-supervised training, eliminating the need to pre-compute exact solutions as ground-truth labels, an often costly or even infeasible requirement for many \ac{CO} problems.
\autoref{fig:approach-overview} illustrates the proposed  \ac{PBGE} framework.

    We evaluate our approach on two well-known combinatorial optimization problems, the \acf{TSP} and the minimum $k$-cut problem.
    Our approach improves the performance of a commonly used approximation algorithm by an order of magnitude in the case of minimum $k$-cut while only minimally increasing the runtime.

    In summary, our contributions include:
    \begin{enumerate}
        \item A method for augmenting existing approximation algorithms for \ac{CO} using \acp{GNN};
        \item A self-supervised method for training a \ac{GNN} for \ac{CO} problems;
        \item A novel gradient estimation scheme, \ac{PBGE}, for backpropagating through \ac{CO} approximation algorithms; and
        \item An extensive experimental evaluation of (1), (2), and (3) on common \ac{CO} problems.
    \end{enumerate}
    The code is part of the supplementary material and will be made public upon acceptance of the paper.

\section{Related work}

    The easiest way of training a model for \ac{CO} is to assume the existence of ground truth solutions to the \ac{CO} problems and train in a supervised fashion.
    \citet{a-note-on-learning-algorithms-for-quadratic-assignment-with-gnns, an-efficient-gcn-technique-for-the-tsp} use a \ac{GNN} to predict an approximate solution as a heatmap, which is then decoded into a feasible solution to the \ac{CO} problem using beam search.
    \citet{pointer-networks} introduce pointer networks, which leverage the fact that many \ac{CO} problems ask to identify a subset or a permutation of the input.
    \citet{neural-algorithmic-reasoning-for-co} follow a neural algorithmic reasoning approach to learn to imitate \ac{CO} solvers.
    \citet{exact-co-with-gcns, nn-steiner} replace components of an existing algorithm that would otherwise be expensive to compute with neural networks.
    \citet{approximation-algorithms-for-co-with-predictions} augment existing algorithms using discrete predictions.
    Finally, \citet{difusco, disco} use graph-based denoising diffusion to generate high-quality solutions.
    However, these supervised approaches aren't applicable to such cases where calculating exact solutions for the training problems is not feasible.

    Several approaches have used \acf{RL} to remove this dependence on a labeled dataset \citep{matnet, dimes, symmetric-replay-training--sample-efficiency-rl-co}.
    A common approach is to formulate the \ac{CO} problem as a Markov decision process \citep{neural-co-with-rl, learning-heuristics-for-the-tsp, solving-np-hard-problems-on-graphs-with-extended-alphago-zero, learning-improvement-heuristics-for-routing-problems}.
    \citet{learning-co-algorithms-over-graphs, attention-learn-to-solve-routing-problems} use the \ac{GNN} autoregressively to predict which node should be added next to the solution set and repeat that process until a valid solution is reached.
    \citet{select-and-optimize, h-tsp, nn-baker} tackle large-scale instances by locally optimizing sub-parts of the instance individually.
    However, \ac{RL} is often sample inefficient and difficult to train due to the high variance of the gradient estimations.

    A common self-supervised approach that does not rely on \ac{RL} is to formulate a surrogate loss consisting of a term that encourages high quality solutions and one that softly enforces the constraints
    \citep{erdos-goes-neural, annealed-training-for-co-on-graphs, can-hybrid-geometric-scattering-networks-solve-max-clique, unsupervised-learning-for-solving-the-tsp, tackling-prevalent-conditions-in-unsupervised-co, gcon}.
    \citet{augment-with-care} train self-supervised using a contrastive loss instead.
    \citet{co-with-physics-inspired-gnns, diffusion-model-framework-for-unsupervised-nco} focus on quadratic and polynomial unconstrained binary optimization, allowing them to formulate self-supervised loss functions for these problem families.
    \citet{gnns-for-maximum-constraint-satisfaction} use an LSTM-based architecture to solve binary maximum constraint satisfaction problems, which many \ac{CO} problems can be formulated as.
    \citet{self-labeling-the-job-scheduling-problem} train a pointer network-based model in a self-supervised fashion.
    \citet{let-the-flows-tell} tackle \ac{CO} using GFlowNets \citep{gflownets}.
    \citet{are-gnns-optimal-approximation-algorithms} propose a \ac{GNN} architecture that can capture message passing algorithms with provable approximation guarantees for a large class of \ac{CO} problems.

    \citet{learning-tsp-requires-rethinking-generalization} compare some of the paradigms introduced in other papers in structured experiments.
    \citet{position--rethinking-post-hoc-search-tsp} argue that the common pattern of training the \ac{GNN} using a surrogate loss, but testing it using a decoder, means that the \ac{GNN} is applied inconsistently, leading to uncertain performance during testing.
    Our approach addresses this problem by incorporating the decoder during training.
    There have been two lines of work on backpropagating through \ac{CO} problems.
    Firstly, if we have a set of optimal solutions given as training data, we can use supervised learning to train the GNN to output adjacency matrices as close as possible to the optimal solutions \citep{elmachtoub2022smart}.
    This is often called ``predict, then optimize''.
    Secondly, there are several methods to backpropagate through a non-differentiable \ac{CO} algorithm, such as \citet{imle, adaptive-imle, differentiation-of-blackbox-co-solvers}.
    Related to our work is decision-focused learning, which has developed several methods to backpropagate through CO solvers \citep{decision-focused-learning}.
    Our approach follows this paradigm.
    The \ac{I-MLE} \citep{imle} is another such method; we compare against it in our experiments and show that our approach improves over \ac{I-MLE}.

\section{Background}
\label{sec:background-co-problems}

    \paragraph{Combinatorial optimization problems.}
    A \acf{CO} problem asks us, given a discrete set $M$ and an objective function $J_\text{CO}: M \rightarrow \mathbb{R}$, to find the minimum
    \[
        \min_{x \in M} J_\text{CO}(x).
    \]
    Maximization problems can be turned into minimization problems by inverting the sign of the objective function. Since finding the exact global optimum is often not required in practice, this paper focuses on efficiently finding approximate solutions.

    To illustrate our approach, we will refer to specific \ac{CO} problems as follows.
    The minimum $k$-cut problem asks, given a weighted, undirected graph and a desired number $k$ of connected components, to find a set of edges with minimum total weight whose removal leaves the graph with exactly $k$ connected components.
    A commonly used approximation algorithm for this problem is the Karger--Stein algorithm \citep{karger-stein}.
    In the \acf{TSP}, we are given a weighted, undirected graph, and are asked to find a minimum-weight Hamiltonian cycle, i.e. a cycle that visits every node exactly once and where the sum of the weights of edges that are traversed in the cycle is minimal.
    A well-known probabilistic approximation algorithm for solving the \ac{TSP} is the random insertion algorithm \citep{random-insertion}.
    The variant of the \ac{TSP} most commonly experimented on in related literature~\citep{attention-learn-to-solve-routing-problems, an-efficient-gcn-technique-for-the-tsp, learning-tsp-requires-rethinking-generalization} is the Euclidean \ac{TSP}, where the graph is fully connected, the nodes represent points in the unit square and the edge weights are the distances between these points.
    We give formal definitions for both problems as well as descriptions of the algorithms mentioned in \autoref{sec:co-problems-and-algorithms}.

    \paragraph{Graph neural networks (GNNs).}
    Let $\mathcal{G} = (V, E)$ be a graph with $n=|V|$ the number of nodes. Let $\mathbf{X} \in \mathbb{R}^{n \times d}$ be the feature matrix that associates each node of the graph with a $d$-dimensional feature vector and let $\mathbf{A} \in \mathbb{R}^{n \times n}$ be the adjacency matrix. GNNs  based on the message passing paradigm have three basic operations, which are defined as
    \begin{equation*}
        \label{eq:mpnn}
        \mathbf{h}_i^{\ell} = \gamma \left(\mathbf{h}_i^{\ell-1}, 
        \varphi_{j \in \mathcal{N}(v_i)} \phi \left(\mathbf{h}_i^{\ell-1}, \mathbf{h}_j^{\ell-1}, r_{ij} \right)\right),
    \end{equation*}
    where $\gamma$, $\varphi$, and $\phi$ represent update, aggregation and message function respectively. \\
    \textbf{Propagation step.} The message-passing network computes a message $m_{ij}^{\ell} = \phi(\mathbf{h}_i^{\ell-1}, \mathbf{h}_j^{\ell-1}, r_{ij})$ between every pair of neighboring nodes $(v_i, v_j)$. The function takes in input $v_i$'s and $v_j$'s representations $\mathbf{h}_i^{\ell-1}$ and $\mathbf{h}_j^{\ell-1}$ at the previous layer $\ell - 1$, and the relation $r_{ij}$ between the two nodes. \\
    \textbf{Aggregation step.} For each node in the graph, the network performs an aggregation computation over the messages from $v_i$'s neighborhood $\mathcal{N}(v_i)$ to calculate an aggregated message $M_i^\ell = \varphi(\{m_{ij}^\ell \mid v_j \in \mathcal{N}(v_i)\})$. The definition of the aggregation function differs between methods. \\
    \textbf{Update step.} Finally, the model non-linearly transforms the aggregated message $M_i^\ell$ and $v_i$'s representation from previous layer $\mathbf{h}_i^{\ell-1}$ to obtain $v_i$'s representation at layer $\ell$ as $\mathbf{h}_i^{\ell} = \gamma(\mathbf{h}_i^{\ell-1}, M_i^\ell)$.

    \paragraph{Gradient Estimators for Discrete Distributions.}
    \label{sec:reinforce}
    Learning systems that backpropagate through discrete distributions and algorithms require gradient estimation approaches. This is because gradients cannot be directly computed or are zero almost everywhere for operations such as sampling from a categorical distribution or running a solver for CO problems. The standard estimator is the score function estimator, also known as the REINFORCE estimator \citep{reinforce}.
    Given a  function $J$ and a parameterized probability distribution $p_\theta(x)$ it estimates the true gradient as follows:
    \[
        \nabla_\theta \mathbb{E}_{y \sim p_\theta(x)} \big[ J(y) \big]
        = \mathbb{E}_{y \sim p_\theta(x)} \big[ J(y) \nabla_\theta \log p_\theta(x) \big]
        \approx \frac{1}{S} \sum_{i=1}^S J(y_i) \nabla_\theta \log p_\theta (x_i), \quad y_i \sim p_\theta(x_i)
    \]
    While this estimator is unbiased, it suffers from large variance. There are other estimators, such as the straight-through estimator \citep{straight-through-estimator} and \ac{I-MLE} \citep{imle} with smaller variance but biased gradient estimates.

\section{Problem statement}
\label{sec:problem-statement}

    We consider \ac{CO} problems on graphs with a linear objective function $J_\text{CO}$ and a probabilistic approximation algorithm $h(\hat{\bm{y}} \mid \gG)$.
    The approximation algorithm takes as input a graph $\gG = (V, E, w)$ with nodes $V$, edges $E$ and edge weights $w: E \rightarrow \mathbb{R}_{> 0}$, and returns (samples) a potentially suboptimal solution $\hat{\bm{y}}$.
    For instance, we might have the minimum $k$-cut problem and the Karger--Stein algorithm.

    We now want to use a GNN $f_{\bm{\theta}}$ parameterized by $\bm{\theta}$ applied to the input graphs $\gG$ to compute an updated graph $\gG' = f_{\bm{\theta}}(\gG)$ such that the probabilistic approximation algorithm when applied to this new graph is improved in expectation.
    Hence, we want to solve the following optimization problem:
    \[
        \min_{\bm{\theta}} \mathbb{E}_{\hat{\bm{y}} \sim h(\hat{\bm{y}} \mid f_{\bm{\theta}}(\gG))} \big[ J_\text{CO}(\hat{\bm{y}}) \big].
    \]
    For each input graph $\gG$, $h(\hat{\bm{y}} \mid f_{\bm{\theta}}(\gG))$ is a discrete probability distribution (due to the assumption that $h$ is probabilistic) parameterized by $\bm{\theta}$.
    The main challenge in solving this optimization problem is that (discrete) approximation algorithms are typically not differentiable functions and that optimal solutions are prohibitively expensive to obtain as training data, making supervised training infeasible.
    Moreover, we assume that the approximation algorithm is a black box---while we can sample from the probability distribution defined by it, we cannot compute a probability mass for a given sample.

\section{Method}
\label{sec:method}

    We introduce a novel approach to \ac{CO} by deriving a new gradient estimator that enables backpropagation through probabilistic approximation algorithms.
    This allows us, for the first time, to directly leverage solution quality rankings, which are automatically generated via objective function-based comparisons, to learn to improve \ac{CO} approximation algorithms.
    We use \acfp{GNN} to guide existing probabilistic approximation algorithms for a given \ac{CO} problem.
    The \ac{GNN} receives the problem graph as input and produces a prior score for each edge.
    These scores are used as additional input alongside the graph for a parameterized version of an off-the-shelf \ac{CO} approximation algorithm, which then produces a solution to the \ac{CO} problem.
    \autoref{fig:approach-overview} shows an overview of our approach.

    Since the approximation algorithm is not differentiable in general, we use gradient estimation to obtain the gradients with respect to the \ac{GNN}'s output.
    Existing gradient estimation schemes, such as REINFORCE, the straight-through estimator, Gumbel softmax, or \ac{I-MLE}, either exhibit high bias or variance or require a differentiable loss function. 
    We propose a new gradient estimation scheme based on preference-based optimization, which we term preference-based gradient estimation (\ac{PBGE}).

    \subsection{Parameterizing approximation algorithms}
    \label{sec:parameterizing-approximation-algorithms}

         An approximation algorithm that takes a problem graph as input can be indirectly parameterized by modifying the graph before executing the algorithm. The modified graph influences the behavior and output of the approximation process.
        We modify an input graph $\gG'$ by using the \acs{GNN}'s output to change its edge weights.
        Assume there is an arbitrary but fixed ordering of edges.
        The model outputs for each edge a prior score $\bm{s} = f_{\bm{\theta}}(\gG) \in \mathbb{R}^{|E|}$.
        A high score for a given edge is interpreted to mean that the respective edge should belong to the solution set with a higher probability mass.
        The approximation algorithms we use prefer including edges of low weight in the solution set. Therefore, we scale down the weights of edges that received high scores.
        Specifically, the edge weights are multiplied with $1 - \sigma(\bm{s})$, where $\sigma$ is the element-wise sigmoid function.
        By running the \ac{CO} approximation algorithm on this modified graph, we parameterize the approximation algorithm using the \acs{GNN}'s output scores $\bm{s}$.
        In the remainder of this paper, $h(\hat{\bm{y}} \mid \gG, \bm{s}) = h(\hat{\bm{y}} \mid \gG')$ denotes the probability distribution defined by a probabilistic \ac{CO} approximation algorithm parameterized in this way.
        It samples and outputs a vector $\hat{\bm{y}} \in \{ 0, 1 \}^{|E|}$ that represents a solution to the \ac{CO} problem, such as a TSP tour or $k$-cut.
        A value of $1$ in $\hat{\bm{y}}$ means that the corresponding edge is in the solution set.

    \subsection{Preference-based gradient estimation (PBGE)}
    \label{sec:preference-based-gradient-estimation}

        In preference learning, a training instance consists of an input and a pair of possible outputs.
        The supervision signal is an annotation indicating that one of the outputs $\bm{y}_w$ is of higher quality than the other output $\bm{y}_l$.
        We can construct a similar setup for \ac{CO} by leveraging a pre-existing probabilistic \ac{CO} approximation algorithm $h$. Sampling from the approximation algorithm multiple times likely yields two solutions $\hat{\bm{y}}_1, \hat{\bm{y}}_2 \sim h(\hat{\bm{y}} \mid \gG, \bm{s})$ of different quality for a given problem instance $\gG \sim \mathcal{D}$ from dataset $\mathcal{D}$.
        These solutions can easily be ranked by applying the \ac{CO} problem's objective function
        \footnote{While $J_\mathrm{CO}$ depends on graph $\gG$ and predicted solution $
        \hat{\bm{y}}$, we omit the graph parameter for readability.}
        $J_\mathrm{CO}$.
        This means assigning $\hat{\bm{y}}_w$ and $\hat{\bm{y}}_l$ such that $J_\mathrm{CO}(\hat{\bm{y}}_w) \leq J_\mathrm{CO}(\hat{\bm{y}}_l)$.

        We now propose the following preference-based loss function:
        \begin{align}
            \begin{split}
                \label{eq:initial-preference-learning-loss}
                \mathcal{L}(\mathcal{D}, \bm{s}) =
                &\; \mathbb{E}_{\hat{\bm{y}}_w, \hat{\bm{y}}_l \sim h(\hat{\bm{y}} \mid \gG, \bm{s}), \: \gG \sim \mathcal{D}} 
                \left[
                    d(\hat{\bm{y}}_w, \hat{\bm{y}}_l)
                    \log \left( \frac{
                        h(\hat{\bm{y}}_l \mid \gG, \bm{s})
                    }{
                        h(\hat{\bm{y}}_w \mid \gG, \bm{s})
                    } \right)
                \right].
            \end{split}
        \end{align}
        Here, $d(\hat{\bm{y}}_w, \hat{\bm{y}}_l)$ is a scaling factor.%
        \footnote{The input graph $\gG$ is also an argument of $d$, but we omit it for ease of readability.}
        As we will see later, its purpose is to scale the gradients based on the distance between the objective values of $\hat{\bm{y}}_w$ and $\hat{\bm{y}}_l$.

        Since we treat the \ac{CO} approximation algorithm as a black box, we cannot calculate the probabilities $h(\hat{\bm{y}}_w \mid \gG, \bm{s})$ and $h(\hat{\bm{y}}_l \mid \gG, \bm{s})$ directly.
        We therefore introduce a proxy distribution $\pi(\hat{\bm{y}} \mid \gG, \bm{s}) \approx h(\hat{\bm{y}} \mid \gG, \bm{s})$ for which we can obtain  probabilities directly.
        For all approximation algorithms used in this paper, a high prior score in $\bm{s}$ for a certain edge increases the probability of this edge being included in the output $\hat{\bm{y}}$.
        This motivates the use of an exponential family distribution to model the proxy distribution $\pi$ for $h$:
        \begin{equation}
            \label{eq:pi-definition}
            \pi(\hat{\bm{y}} \mid \gG, \bm{s}) = \frac{\exp(\langle \hat{\bm{y}}, \bm{s} \rangle)}{\sum_{\bm{y}' \in \mathcal{C}} \exp(\langle \bm{y}', \bm{s} \rangle)},
        \end{equation}
        where $\langle \cdot, \cdot \rangle$ is the inner product and $\mathcal{C}$ is the set of all solutions to the \ac{CO} problem.

        Replacing $h$ with $\pi$ in \autoref{eq:initial-preference-learning-loss} and inserting \autoref{eq:pi-definition} simplifies the loss function to
        \begin{align}
            \begin{split}
                \mathcal{L}(\mathcal{D}, \bm{s}) =
                &\; \mathbb{E}_{\bm{\hat{y}}_w, \bm{\hat{y}}_l \sim h(\hat{\bm{y}} \mid \gG, \bm{s}), \: \gG \sim D} \Big[
                    d(\hat{\bm{y}}_w, \hat{\bm{y}}_l)
                    \big( \langle \bm{\hat{y}}_l, \bm{s} \rangle - \langle \bm{\hat{y}}_w, \bm{s} \rangle \big)
                \Big].
            \end{split}
        \end{align}
        Now, the gradient of this expectation with respect to $\bm{s}$ is 
        \begin{align}
            \begin{split}
                \label{eq:loss-derivative}
                \nabla_{\bm{s}} \mathcal{L}(\mathcal{D}, \bm{s}) =
                &\; \mathbb{E}_{\bm{\hat{y}}_w, \bm{\hat{y}}_l \sim h(\hat{\bm{y}} \mid \gG, \bm{s}), \: \gG \sim D} \big[
                    d(\hat{\bm{y}}_w, \hat{\bm{y}}_l)
                    (\bm{\hat{y}}_l - \bm{\hat{y}}_w)
                \big],
            \end{split}
        \end{align}
        whose single-sample Monte Carlo estimate can be written as
        \begin{align}
            \begin{split}
                \nabla_{\bm{s}} \mathcal{L}(\gG, \bm{s})
                &\approx d(\hat{\bm{y}}_w, \hat{\bm{y}}_l) (\hat{\bm{y}}_l - \hat{\bm{y}}_w), \quad \text{where } \hat{\bm{y}}_w, \hat{\bm{y}}_l \sim h(\hat{\bm{y}} \mid \gG, \bm{s}).
            \end{split}
        \end{align}
        Intuitively, the gradient is negative at a certain edge if that edge is in the better solution, but not in the worse solution.
        A negative gradient raises the \ac{GNN}'s output score, meaning that the \ac{GNN} will be nudged towards including this edge in its solution.
        Similarly, a positive gradient means that the corresponding edge was in the worse solution, but not in the better solution.
        This nudges the \ac{GNN}'s output down, so it pushes the \ac{GNN} towards not including this edge.

        The form of the gradient is reminiscent of those used for preference learning with \acp{LLM} \citep{direct-preference-optimization, meng2024simpo}. There, the gradient makes the model increase the likelihood of the better solution and decrease the likelihood of the worse solution, and a scaling factor is used to weight important gradients more highly.
        Unlike the preference learning setting used with \acp{LLM}, we not only know which solution in a pair is better, but we can measure the quality of each solution exactly using the objective function.
        This eliminates the need for human annotators to rank pairs of examples.
        Moreover, we can leverage the objective function to more easily compute a suitable scaling factor.

        \paragraph{Gradient scaling.}
        If the solutions $\hat{\bm{y}}_w$ and $\hat{\bm{y}}_l$ are of similar quality, we do not want to strongly move the \ac{GNN} towards either solution.
        We therefore scale the gradient with the relative optimality gap between the two solutions,
        \[
            d(\hat{\bm{y}}_w, \hat{\bm{y}}_l) = \frac{J_\text{CO}(\hat{\bm{y}}_l)}{J_\text{CO}(\hat{\bm{y}}_w)} - 1.
        \]
        This is always non-negative, since, by definition, $J_\text{CO}(\hat{\bm{y}}_l) \geq J_\text{CO}(\hat{\bm{y}}_w)$.
        Using this scaling factor places more weight on pairs of solutions where the difference in their respective optimality gaps is large.
        In particular, if the two solutions are of the same quality, the gradient is set to zero, so we do not move the \ac{GNN} towards either solution.
        The scaled gradient is
        \[
            \nabla_{\bm{s}} \mathcal{L}(\gG, \bm{s})
            \approx \left( \frac{J_\text{CO}(\hat{\bm{y}}_l)}{J_\text{CO}(\hat{\bm{y}}_w)} - 1 \right)
            \big( \hat{\bm{y}}_l - \hat{\bm{y}}_w \big).
        \]
        
        \begin{wrapfigure}[7]{R}{0.51\textwidth}
            \vspace{-4.5ex}
            \small
            \hfill
            \begin{minipage}{0.505\textwidth}
                \begin{algorithm}[H]
                    \caption{One training iteration with \ac{PBGE}}
                    \label{alg:training-with-pbge}
                    \begin{algorithmic}
                        \State $\bm{s} \gets f_{\bm{\theta}}(\gG)$
                        \smallskip
                        \State \Comment{Sample $n$ solutions from $h$ guided by $\bm{s}$, and $m$ unguided solutions}
                        \State $\hat{\bm{y}}_1, \dots, \hat{\bm{y}}_n \sim h(\hat{\bm{y}} \mid \gG, \bm{s})$
                        \State $\hat{\bm{y}}_{n+1}, \dots, \hat{\bm{y}}_{n+m} \sim h(\hat{\bm{y}} \mid \gG)$
                        \State $\hat{\mathcal{Y}} \gets \{\hat{\bm{y}}_1, \dots, \hat{\bm{y}}_{n+m}\}$
                        \State $\hat{\bm{y}}_w \gets \argmin_{\hat{\bm{y}} \in \hat{\mathcal{Y}}} J_\mathrm{CO}(\hat{\bm{y}})$
                        \smallskip
                        \State \Comment{Estimate gradient using each pair involving $\hat{\bm{y}}_w$}
                        \State $\nabla_{\bm{s}} \mathcal{L}(\gG, \bm{s}) \approx
                                \displaystyle\sum_{\hat{\bm{y}}_l \in \hat{\mathcal{Y}}}
                                \! \left( \frac{J_\text{CO}(\hat{\bm{y}}_l)}{J_\text{CO}(\hat{\bm{y}}_w)} - 1 \right)
                                \! \big( \hat{\bm{y}}_l - \hat{\bm{y}}_w \big)$
                        \State Backpropagate gradient $\nabla_{\bm{s}} \mathcal{L}(\gG, \bm{s})$
                    \end{algorithmic}
                \end{algorithm}
            \end{minipage}
        \end{wrapfigure}

        The variance of the gradient can be reduced by estimating the expectation in \autoref{eq:loss-derivative} by creating a pool of solutions with the approximation algorithm and by constructing pairs from this pool.
        We form the pairs by combining the best solution from the pool with each of the weaker solutions.
        In practice, the accuracy of the gradients depends heavily on the quality of the best found solution $\hat{\bm{y}}_w$.
        At the beginning of training, the \ac{GNN} cannot yet output good enough scores to consistently find reasonable $\hat{\bm{y}}_w$.
        To remedy this, we also run the approximation algorithm on the unmodified graph and add the resulting solutions to the pool from which the pairs are generated.
        The complete training procedure is described in Algorithm~\ref{alg:training-with-pbge}.

        At test time, the model's output needs to be converted (decoded) to a solution to the \ac{CO} problem.
        This can simply be done by running the \ac{CO} approximation algorithm with the model's output as input, as described in \autoref{sec:problem-statement}.
        The solution can be improved by running a probabilistic \ac{CO} approximation algorithm repeatedly and using the best solution found as final output.

\newpage
    \subsection{Theoretical analysis}

        We prove the following theoretical results, which characterize under what conditions we can turn an approximation algorithm into an exact algorithm if we find an optimal modified input graph. First, we prove this for the Karger--Stein algorithm.
        \begin{tcolorbox}
            \begin{theorem}
                Let $h(\hat{\bm{y}} \mid \gG, \bm{s})$ be the Karger--Stein algorithm guided by scores $\bm{s} \in \mathbb{R}^{|E|}$, and let $\bm{y} \in \{0, 1\}^{|E|}$ be a minimum $k$-cut on graph $\gG = (V, E)$. Then,
                \[
                    \lim_{\sigma(\bm{s}) \to \bm{y}} h(\bm{y} \mid \gG, \bm{s}) = 1.
                \]
            \end{theorem}
        \end{tcolorbox}
        Now, we consider an arbitrary \ac{CO} problem whose solutions are a set of edges of minimum total weight that satisfy the constraints (e.g., \ minimum $k$-cut or \ac{TSP}).
        \begin{tcolorbox}
            \begin{theorem}
                Let $J_\text{CO}$ be the objective of a minimization problem on graphs with  $J_\text{CO}(\hat{\bm{y}}) = \sum_{e \in \hat{\bm{y}}} w(e)$ and $w(e) > 0$ and let $\bm{y} \in \{0, 1\}^{|E|}$ be an optimal solution on graph $\gG$.
                Moreover, let $h(\hat{\bm{y}} \mid \gG)$ be a probabilistic approximation algorithm with a bound on the relative error, that is, there exists a function $\alpha$ such that for all $\hat{\bm{y}} \sim h(\hat{\bm{y}} \mid \gG)$, $J_\text{CO}(\hat{\bm{y}}) \!\leq\! \alpha(\gG) J_\text{CO}(\bm{y})$.
                Then,
                \[
                    \lim_{\sigma(\bm{s}) \to \bm{y}} h(\bm{y} \mid \gG, \bm{s}) = 1.
                \]
            \end{theorem}
        \end{tcolorbox}
        In other words, in the limit, the approximation algorithm guided by $\bm{s}$ always finds optimal solution $\bm{y}$.
        From this, the following result regarding insertion algorithms on \ac{TSP} immediately follows:

        \begin{corollary}
            Let $\gG = (V, E, w)$ be an undirected graph, and let $h(\hat{\bm{y}} \mid \gG)$ be a probabilistic insertion algorithm, such as random insertion.
            Let $\bm{y} \in \{0, 1\}^{|E|}$ be a minimal length TSP tour on $\gG$.
            Then,
            \[
                \lim_{\sigma(\bm{s}) \to \bm{y}} h(\bm{y} \mid \gG, \bm{s}) = 1.
            \]
        \end{corollary}

        More rigorous formulations of these theorems and their proofs can be found in \autoref{sec:proofs}.

\section{Experiments}

    We validate our approach on two well-known combinatorial optimization problems: the \acf{TSP} and the minimum $k$-cut problem.
    For both problems, we synthetically generate problem instances and establish baselines as reference.
    We use residual gated graph convnets \citep{residual-gated-graph-convnets}, but adapt them to include edge features $\bm{e}_{ij}^l$ and a dense attention map $\bm{\eta}_{ij}^l$ following \citet{an-efficient-gcn-technique-for-the-tsp}.
    Please see \autoref{sec:implementation-details} for details.

        \begin{table*}
            \centering
            \caption{
                Minimum $k$-cut optimality gaps on graphs with 100 nodes, using Karger--Stein as decoder.
                Mean $\pm$ standard deviation were calculated over ten evaluation runs on the same model parameters.
                In the supervised and self-supervised rows, a \ac{GNN} trained with the indicated method is used to guide the Karger--Stein algorithm.
                In the columns labeled ``Best out of 3 runs'', the Karger--Stein algorithm is run three times on the same GNN outputs, and the best result is used.
            }
            \label{tab:minimum-k-cut-optimality-gaps}
            \resizebox{\textwidth}{!}{%
            \begin{tabular}{lcc@{\hskip 2em}cc}
                \toprule
                \textbf{Method}                   & \multicolumn{2}{c}{\textbf{Unweighted graphs} \hspace*{2em}} & \multicolumn{2}{c}{\textbf{NOIgen+} \hspace*{0.5em}} \\
                                                  & Single run & Best out of 3 runs & Single run & Best out of 3 runs \\
                \midrule
                \textbf{Non-learned}              &              &                                 &                                 \\
                \quad Karger--Stein               & 3.61\%$\pm$0.21 (330ms) & 0.62\%$\pm$0.07 (971ms) & 11.29\%$\pm$0.71 (352ms) & 0.43\%$\pm$0.11 (1.00s) \\
                \midrule
                \textbf{Supervised}                &         &                &                &                \\
                \quad \acs{BCE} loss               & 0.27\%$\pm$0.06 (402ms) & 0.03\%$\pm$0.02 (1.05s) & 0.41\%$\pm$0.07 (415ms) & 0.06\%$\pm$0.04 (1.04s) \\
                \quad \ac{I-MLE}                   & 1.67\%$\pm$0.10 (415ms) & 0.14\%$\pm$0.06 (1.03s) & 2.53\%$\pm$0.12 (435ms) & 0.28\%$\pm$0.06 (1.08s) \\
                \midrule
                \textbf{Self-supervised}           &     &                &                &                \\
                \quad \ac{I-MLE}                   & 3.39\%$\pm$0.16 (402ms) & 0.52\%$\pm$0.05 (1.01s) & 7.63\%$\pm$0.41 (444ms) & 0.41\%$\pm$0.07 (1.10s) \\
                \quad PBGE (\textbf{ours})         & 0.38\%$\pm$0.05 (398ms) & 0.06\%$\pm$0.05 (1.01s) & 0.58\%$\pm$0.06 (439ms) & 0.09\%$\pm$0.05 (1.08s) \\
                \bottomrule
            \end{tabular}}
        \end{table*}

    \subsection{Problem instance generation and baselines}
    \label{sec:graph-generation}

        For minimum $k$-cut, we use the established graph generator \emph{NOIgen}~\citep{noigen}.
        Since it relies on dramatically scaling down the weights of edges that are in the minimum $k$-cut in order to avoid trivial solutions, it makes it easy for a \ac{GNN} to identify the correct edges.
        To make the graphs more challenging, we extend NOIgen to also use graph structure to avoid trivial solutions, which allows us to scale down edge weights less dramatically.
        We call this improved graph generator \emph{NOIgen+}.
        We also generate unweighted graphs that only rely on graph structure to prevent trivial solutions.
        Graphs for the \ac{TSP} are generated according to the established method described in \citet{attention-learn-to-solve-routing-problems, an-efficient-gcn-technique-for-the-tsp, learning-tsp-requires-rethinking-generalization}.
        Please refer to \autoref{sec:graph-generation-appendix} for further details.

        \begin{figure}[t!]
            \vspace{-0.2ex}
            \begin{subfigure}{0.49\linewidth}
                \includegraphics[width=\textwidth]{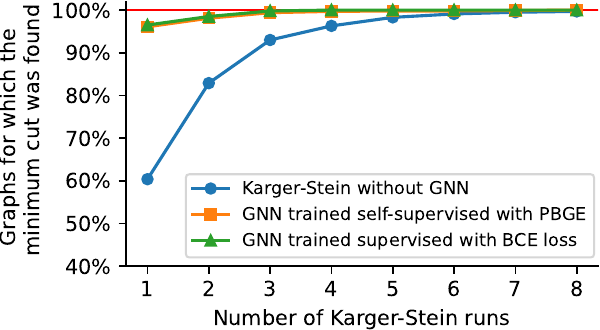}
                \caption{On graphs without edge weights and 100 nodes.}
                \label{fig:karger-stein-num-runs-until-minimum-found-n-100}
            \end{subfigure}
            \hfill
            \begin{subfigure}{0.49\linewidth}
                \includegraphics[width=\textwidth]{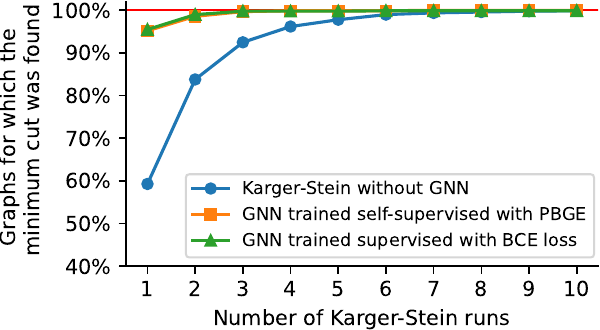}
                \caption{On graphs without edge weights and 200 nodes.}
                \label{fig:karger-stein-num-runs-until-minimum-found-generalising-to-n-200}
            \end{subfigure}

            \caption{
                The number of graphs for which the minimum $k$-cut was found after a given number of Karger--Stein runs.
                For example, for Karger--Stein without a \ac{GNN} on graphs with 100 nodes (\autoref{fig:karger-stein-num-runs-until-minimum-found-n-100}), two Karger--Stein runs suffice to find the minimum $k$-cut for 83\% of graphs.
            }
            \label{fig:karger-stein-num-runs-until-minimum-found}
        \end{figure}

        Since we assume a setting without access to ground truth solutions to the \ac{CO} problems, our primary baselines are gradient estimation schemes for unsupervised training.
        We set the loss to
        \[
            \mathcal{L}(\mathcal{D}, \bm{s}) = \mathbb{E}_{\hat{\bm{y}} \sim h(\hat{\bm{y}} \mid \gG, \bm{s}), \: \gG \sim D}
            \big[ J_\mathrm{CO}(\hat{\bm{y}}) \big]
        \]
        and estimate $\nabla_{\bm{s}} \mathcal{L}(\mathcal{D}, \bm{s})$ using \ac{I-MLE} \citep{imle}, and, in the case of \ac{TSP}, REINFORCE \citep{reinforce}.
        Here, I-MLE acts as a representative of the decision-focused learning paradigm \citep{decision-focused-learning}.

        For minimum $k$-cut, we also train supervised baselines using an edge-level \ac{BCE} loss comparing the \acs{GNN}'s output scores $\bm{s}$ with a pre-calculated ground truth solution $\bm{y}$.
        Additionally, we train models using \ac{I-MLE} in a supervised fashion, comparing the approximation algorithm's output $\hat{\bm{y}}$ with $\bm{y}$ using a Hamming loss.
        See \autoref{sec:baseline-details} for details on these baselines.

    \subsection{Results}

        \paragraph{Minimum \texorpdfstring{$k$}{k}-Cut.}
        We evaluate our method on the minimum $k$-cut problem, using the Karger--Stein algorithm as a base.
        \autoref{tab:minimum-k-cut-optimality-gaps} shows optimality gaps of the unmodified Karger--Stein algorithm, as well as several versions of our method.
        Each version augments the Karger--Stein algorithm with a \ac{GNN}, and they differ by how the \ac{GNN} was trained.
        Note that when augmenting the Karger--Stein algorithm with a \ac{GNN} trained with PBGE, the optimality gap improves by an order of magnitude.
        On top of this, even though it didn't use any ground truth solutions during training, the \ac{GNN} trained with \ac{PBGE} comes close to matching the \ac{GNN} trained supervised with a \ac{BCE} loss.

        In practice, the most important metric is the number of runs for Karger--Stein to find the optimal $k$-cut.
        If this number is low, we can run Karger--Stein a small number of times and be reasonably certain that the minimum $k$-cut was found.
        \autoref{fig:karger-stein-num-runs-until-minimum-found} shows for how many graphs the minimum $k$-cut is found in a set number of runs, comparing the unmodified Karger--Stein algorithm with two versions that were augmented using a \ac{GNN}.
        On both datasets, the augmented Karger--Stein algorithm needs much fewer runs to find the minimum $k$-cut, almost always finding it on the first attempt.
        Again, the \ac{GNN} trained self-supervised with PBGE comes close to matching supervised performance.

        \begin{table*}[t!]
            \centering
            \caption{
                TSP optimality gaps, with mean $\pm$ standard deviation, calculated over ten evaluation runs on the same model parameters.
                We indicate the decoder used in parentheses.
                Values with * are obtained from these papers and therefore do not include standard deviations.
                We provide an extended version of this table with more baselines in \autoref{sec:extended-comparison}.
            }
            \label{tab:tsp-optimality-gaps}
            \resizebox{\textwidth}{!}{%
            \begin{tabular}{llll}
                \toprule
                \textbf{Method} (decoder in parentheses)   & $\bm{n = 20}$   & $\bm{n = 50}$   & $\bm{n = 100}$  \\
                \midrule
                \textbf{Self-supervised}                   &                 &                 &                 \\
                \citeauthor{neural-co-with-rl} (greedy)          & 1.42\%*        & 4.46\%*          & 6.90\%*          \\
                \citeauthor{learning-co-algorithms-over-graphs} (greedy) & 1.42\%*         & 5.16\%*         & 7.03\%*         \\
                \citeauthor{learning-heuristics-for-the-tsp} (greedy) & 0.66\%* (2m) & 3.98\%* (5m) & 8.41\%* (8m) \\
                \citeauthor{learning-heuristics-for-the-tsp} (greedy + 2OPT) & 0.42\%* (4m) & 2.77\%* (26m) & 5.21\%* (3h) \\
                \citeauthor{learning-heuristics-for-the-tsp} (sampling 1280 times) & 0.11\%* (5m) & 1.28\%* (17m) & 12.70\%* (56m) \\
                \citeauthor{learning-heuristics-for-the-tsp} (sampling 1280 times + 2OPT) & 0.09\%* (6m) & 1.00\%* (32m) & 4.64\%* (5h) \\
                \citeauthor{attention-learn-to-solve-routing-problems} (greedy) & 0.34\%*         & 1.76\%* (2s)    & 4.53\%* (6s)    \\
                \citeauthor{attention-learn-to-solve-routing-problems} (sampling 1280 times) & 0.08\%* (5m)    & 0.52\%* (24m)   & 2.26\%* (1h)    \\
                REINFORCE (random ins., 20 runs)       & 7.98\%$\pm$0.08 (769ms)       & 23.84\%$\pm$0.10 (11.62s) & 52.83\%$\pm$0.47 (1.38m) \\
                REINFORCE (random ins., 100 runs)      & 4.78\%$\pm$0.03 (3.64s)       & 11.24\%$\pm$0.08 (57.85s) & 31.06\%$\pm$0.25 (7.53m) \\
                I-MLE (random ins., 20 runs)           & 10.54\%$\pm$0.15 (761ms)      & 35.87\%$\pm$0.34 (11.14s) & 61.42\%$\pm$0.23 (1.37m) \\
                I-MLE (random ins., 100 runs)          & 6.55\%$\pm$0.10 (3.68s)       & 19.40\%$\pm$0.24 (56.63s) & 44.42\%$\pm$0.14 (7.51m) \\
                PBGE \textbf{(ours)} (random ins., 20 runs)  & 0.18\%$\pm$0.01 (763ms) & 2.37\%$\pm$0.02 (11.11s)  & 5.12\%$\pm$0.06 (1.43m) \\
                PBGE \textbf{(ours)} (random ins., 100 runs) & 0.05\%$\pm$0.01 (3.73s) & 1.13\%$\pm$0.03 (53.98s)  & 3.67\%$\pm$0.05 (7.44m) \\
                \midrule
                \multicolumn{2}{l}{\textbf{Non-learned approximation algorithms}}   &                 &                 \\
                Christofides                               & 8.72\%$\pm$0.00 (45ms) & 11.07\%$\pm$0.00 (685ms) & 11.86\%$\pm$0.00 (4.45s) \\
                Random Insertion                           & 4.46\%$\pm$0.08 (41ms) & 7.57\%$\pm$0.08 (575ms)  & 9.63\%$\pm$0.11 (4.34s) \\
                Farthest Insertion                         & 2.38\%$\pm$0.00 (57ms) & 5.50\%$\pm$0.00 (909ms)  & 7.58\%$\pm$0.00 (5.68s) \\
                LKH3                                       & 0.00\%* (18s)          & 0.00\%* (5m)             & 0.00\%* (21m)   \\
                \bottomrule
            \end{tabular}}
        \end{table*}

        \begin{wrapfigure}[19]{R}{0.52\textwidth}
            \hfill
            \begin{minipage}[T]{0.5\textwidth}
                \includegraphics[width=0.96\linewidth]{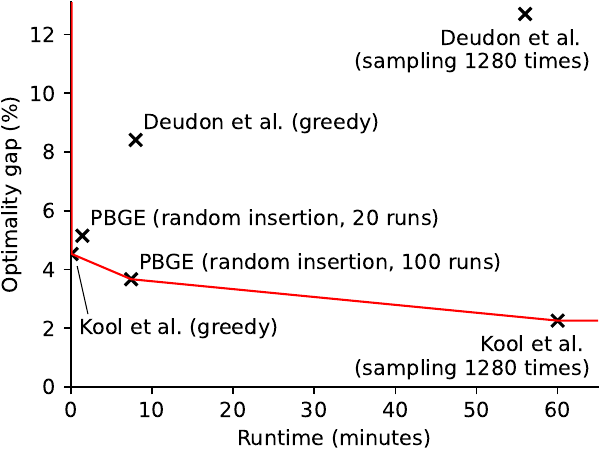}
                \caption{The Pareto frontier of self-supervised methods on TSP-100. Non-Pareto-optimal baselines with very high optimality gaps or very long runtimes are omitted for clarity.}
                \label{fig:tsp-pareto-optimality}
            \end{minipage}
        \end{wrapfigure}

        \paragraph{Travelling salesman problem (TSP).}        
        \autoref{tab:tsp-optimality-gaps} shows the optimality gaps of our approach and its variants on \ac{TSP}.
        All of our models were trained using random insertion as the \ac{CO} approximation algorithm.
        For \ac{PBGE}, we sampled 10 solutions from $h(\hat{\bm{y}} \mid \gG, \bm{s})$ and 10 solutions from $h(\hat{\bm{y}} \mid \gG)$.

        The decoder used at test time is listed after the name of the respective method in parentheses.
        Greedy search starts from an arbitrary node, and follows the edge with the highest score to select the next node.
        This process of greedily following the best edge is repeated until each node has been visited once.
        To ensure that the resulting tour is valid, edges that lead to nodes that have already been visited are excluded.
        Sampling simply refers to sampling multiple solutions and using the best one.
        ``+ 2OPT'' refers to improving the decoded solution using 2OPT local search \citep{2opt}.

        \autoref{fig:tsp-pareto-optimality} shows the Pareto frontier of self-supervised methods on TSP-100.
        \ac{PBGE} achieves Pareto-optimality when decoding with 100 random insertion runs.

\section{Conclusion}
\label{sec:conclusion}

    We introduced a method to improve existing approximation algorithms for \ac{CO} using \ac{GNN}s.
    The \ac{GNN} predicts parameters, which are used as input for the non-learned approximation algorithm to produce a high-quality solution to the \ac{CO} problem.
    The GNN is trained based on the \ac{CO} problem's downstream objective, without the need for labelled data.
    To achieve this, we used gradient estimation to backpropagate through the approximation algorithm.
    We proposed a novel gradient estimation scheme for this purpose, which we called \acf{PBGE}.

    \paragraph{Limitations and future work.}
    Incorporating a \ac{CO} approximation algorithm during training means that the training process is more computationally intensive compared to competing approaches.
    This also means that an existing approximation algorithm is required for our approach.
    We only experimented on \ac{CO} problems for which solutions can be represented in terms of the graph's edges.
    While extending our approach to other kinds of \ac{CO} problems is theoretically possible, we leave this for future work.
    We also plan to incorporate the ability to formulate additional constraints.

\begin{ack}
    The authors would like to thank Aneesh Barthakur, Matteo Palmas, Roman Freiberg, and Jiaqi Wang for fruitful discussions.
    The authors thank the International Max Planck Research School for Intelligent Systems (IMPRS-IS) for supporting Arman Mielke.
    The contributions of Thilo Strauss were carried out while employed with ETAS Research.
\end{ack}

\bibliography{paper}
\bibliographystyle{icml2025}

\newpage
\appendix

\section{Combinatorial optimization problems and approximation algorithms}
\label{sec:co-problems-and-algorithms}

    \subsection{Minimum \texorpdfstring{$k$}{k}-cut problem}
    \label{sec:algorithms-minimum-k-cut}

        \subsubsection{Problem Definition}

            We are given a connected, undirected graph $\gG = (V, E, w)$ with edge weights $w: E \rightarrow \mathbb{R}_{> 0}$, as well as a desired number of connected components $k \in \mathbb{N}, 2 \leq k \leq |V|$.
            The goal is to find a set of edges $C \subseteq E$ with minimal total weight whose removal leaves $k$ connected components.
            This set is called a \emph{minimum $k$-cut}.
            Formally, we are optimizing
            \[
                \min_{C \subseteq E} \sum_{e \in C} w(e) \quad
                \text{ such that graph } (V, E \backslash C) \text{ has } k \text{ connected components}.
            \]

        \subsubsection{Karger's algorithm}
            Karger's algorithm~\citep{karger} is a Monte Carlo algorithm for the minimum $k$-cut problem.

            The algorithm is based on the \emph{contraction operation}:
            An edge $e = \{ x, y \}$ is contracted by merging its nodes $x$ and $y$ into a new node $xy$.
            For clarity, we will call the node that results from this merger a meta-node.
            Every edge that was incident to exactly one of the two merged nodes is now altered to instead be incident to the meta-node $xy$:
            An edge $\{ x, z \}$ or $\{ y, z \}$ becomes $\{ xy, z \}$.
            This may result in parallel edges, meaning that the resulting graph is a multigraph.
            All edges $\{ x, y \}$ are removed, so that the resulting multigraph contains no self-loops.

            Karger's algorithm works by repeatedly sampling an edge, where the probability of each edge is proportional to its weight, then contracting that edge.
            This is repeated until there are only $k$ nodes left.
            Each of these remaining $k$ meta-nodes represents a connected component in the original graph, with each node of the original graph that was subsequently been merged into that meta-node belonging to this connected component.
            Any edge in the original graph that spans between two connected components is cut.

            Since this algorithm is not guaranteed to find the minimum $k$-cut, a common strategy is to run the algorithm repeatedly and use the smallest found cut as the final result.

        \subsubsection{Karger--Stein algorithm}

            The Karger--Stein algorithm~\citep{karger-stein} is a recursive version of Karger's algorithm, shown in \autoref{alg:karger-stein}.

            \begin{algorithm}[h!]
                \caption{Karger--Stein algorithm}
                \label{alg:karger-stein}
                \begin{algorithmic}
                    \State \textsc{Karger--Stein}
                    \State \textbf{Input:} connected, undirected graph $\gG = (V, E, w)$
                    \If{$|V| \leq 6$}
                        \State \textbf{return} $\textsc{Contract}(\gG, 2)$
                    \Else
                        \State target $t \gets \left\lceil \displaystyle\frac{|V|}{\sqrt{2}} + 1 \right\rceil$
                        \State $\gG_1 \gets \textsc{Contract}(\gG, t)$
                        \State $\gG_2 \gets \textsc{Contract}(\gG, t)$
                        \State \textbf{return} $\min \! \big\{ \textsc{Karger--Stein}(\gG_1), \; \textsc{Karger--Stein}(\gG_2) \big\}$ \Comment{Return the lower-weight cut}
                    \EndIf
                    \bigskip
                    \State \textsc{Contract}
                    \State \textbf{Input:} connected, undirected graph $\gG = (V, E, w)$, target number of nodes $t$
                    \While{$|V| > t$}
                        \State $\gG \gets$ sample edge in $\gG$ and contract it
                    \EndWhile
                    \State \textbf{return} $\gG$
                \end{algorithmic}
            \end{algorithm}

    \subsection{Travelling Salesman Problem (TSP)}

        \subsubsection{Problem Definition}

            We are given a complete directed or undirected graph $\gG = (V, E, w)$ with edge weights $w: E \longrightarrow \mathbb{R}$.
            Our goal is to find a minimum weight Hamiltonian cycle in $\gG$.
            A Hamiltonian cycle, also called a tour, is a sequence of nodes where each node in the graph appears exactly once, each node in the sequence is adjacent to the previous node, and the first node is adjacent to the last node.
            The weight of the cycle is the sum of all edge weights of edges between nodes that appear next to each other in the cycle, including the edge from the last to the first node.
            Instead of representing a tour as a sequence of nodes, we will represent it as the set of edges between nodes that appear next to each other in the sequence.
            Formally, we are optimizing
            \[
                \min_{T \subseteq E} \sum_{e \in T} w(e) \quad
                \text{ such that } T \text{ forms a Hamiltonian cycle}.
            \]

            The nodes are commonly called cities, and the weight of an edge is commonly called the distance between the two cities.
            The TSP is NP complete.
    
            \paragraph{Metric TSP.}
            The metric version of TSP additionally assumes that the distances between the cities form a metric.
            This means that
            \begin{itemize}
                \item the graph is undirected (or $w(x, y) = w(y, x)$ for all $x, y \in V$),
                \item distances between cities at different locations are positive, and
                \item the edge weights satisfy the triangle inequality, i.e.\ $w(x, y) + w(y, z) \geq w(x, z)$ for all $x, y, z \in V$.
            \end{itemize}
    
            \paragraph{Euclidean TSP.}
            Euclidean TSP is a special case of metric TSP in which the cities are located at points in the unit square and the distances between the cities are the Euclidean distances between the respective points.
            Research on solving the TSP using neural networks often focuses on this version of the problem \citep{attention-learn-to-solve-routing-problems, an-efficient-gcn-technique-for-the-tsp, learning-tsp-requires-rethinking-generalization}.

        \subsubsection{Random insertion algorithm}
        \label{sec:algorithms-random-insertion}
    
            The random insertion algorithm~\citep{random-insertion} is a Monte Carlo algorithm for the \ac{TSP}.
            
            Since a Hamiltonian cycles to a given graph $\gG = (V, E)$ is required to contain every $v \in V$ exactly once, it is a straightforward approach to iteratively sample and remove nodes from $V$ until it is empty.
            The random insertion algorithm, as suggested by Karg and Thompson, begins by selecting two nodes $s,t \in V$ at random and adds the edges $(s,t)$ and $(t,s)$ to an initial cycle.
            In order to extend the cycle to include all nodes, the algorithm now samples a node $v \in V \setminus \{s,t\}$ and selects the edges $(x,v)$ and $(v,y)$ such that $x$ and $y$ are already part of the partial cycle with $x \neq y$ and such that the sum of the metric distances of $(x,v)$ and $(v,y)$ is minimal.
            
            The cycles obtained in this way are at most $(\lceil \log_2 |V| \rceil + 1)$ times longer than the optimal cycle \citep{insertion-algorithms-approximation-ratio}.
    
            The algorithm is summarized in \autoref{alg:random-insertion}.
    
            \begin{algorithm}[h]
                \caption{Random insertion}
                \label{alg:random-insertion}
                \begin{algorithmic}
                    \State \textbf{Input:} connected, undirected graph $\gG = (V, E, w)$
                    \State $\mathcal{T} \gets$ \ac{TSP} tour consisting of one random node
                    \State $v \gets$ sample node in $V$ that is not part of $\mathcal{T}$ yet
                    \State $\mathcal{T} \gets$ insert $v$ into $\mathcal{T}$ to form a loop of two nodes
    
                    \For{$i \in \{1, \ldots, |V|\}$}
                        \State $v \gets$ sample node in $V$ that is not part of $\mathcal{T}$ yet
                        \State $\mathcal{T} \gets$ insert $v$ into $\mathcal{T}$ at the point in the tour $\mathcal{T}$ where it increases the tour's length by the least amount
                    \EndFor
                    \State \textbf{return} $\gG$
                \end{algorithmic}
            \end{algorithm}

\section{Proofs}
\label{sec:proofs}

    \setcounter{theorem}{0}

    We assume that the edges $E$ of a graph $\gG = (V, E, w)$ are in an arbitrary but fixed order.
    This means that the scores assigned to the edges by a \ac{GNN} can be represented as a vector $\bm{s} \in \mathbb{R}^{|E|}$.
    We use $\bm{s}[e]$ to denote the score assigned to a specific edge $e \in E$.
    A subset of edges $\hat{\bm{y}}$ can be represented as $\hat{\bm{y}} \in \{0, 1\}^{|E|}$, where a 1 indicates that the respective edge is in the set.
    For notational simplicity, we will still write $e \in \hat{\bm{y}}$ for edges that are in this subset.

    A probabilistic \ac{CO} approximation algorithm for an edge subset problem defines a probability distribution over the subsets of edges.
    We use $h(\hat{\bm{y}} \mid \gG, \bm{s})$ to denote the probability that the output is edge subset $\hat{\bm{y}}$ given input graph $\gG$ and edge scores $\bm{s}$, as described in \autoref{sec:method}.

    We use $\sigma$ to denote the element-wise sigmoid function $\sigma(x) = \frac{e^x}{1 + e^x}$.

    \subsection{Probability of finding the optimal minimum \texorpdfstring{$k$}{k}-cut}

        Let $\gG = (V, E)$ be an undirected graph.
        During the first $i - 1$ iterations of the modified Karger's algorithm, some edges have been merged away.
        Let $E_i$ be the set of edges that are left at iteration $i$ (this means $E_1 = E$).
        The probability that a specific edge $e$ is selected for contraction at iteration $i$ is
        \[
            p_i(e) = \frac{1 - \sigma(\bm{s}[e])}{\sum_{e' \in E_i} \big( 1 - \sigma(\bm{s}[e']) \big)}.
        \]

        \begin{tcolorbox}
            \begin{theorem}
                Let $\gG = (V, E)$ be an undirected graph, and let $\bm{y} \in \{0, 1\}^{|E|}$ be a minimum $k$-cut on $\gG$.
                Moreover, let $h(\hat{\bm{y}} \mid \gG, \bm{s})$ be Karger's algorithm guided by parameters $\bm{s} \in \mathbb{R}^{|E|}$, and let $\sigma$ be the element-wise sigmoid function.
                Then,
                \[
                    \lim_{\sigma(\bm{s}) \to \bm{y}} h(\bm{y} \mid \gG, \bm{s}) = 1.
                \]
            \end{theorem}
        \end{tcolorbox}
        In other words, in the limit, Karger's algorithm guided by $\bm{s}$ always finds optimal solution $\bm{y}$.

        \begin{proof}
            Assuming that no edge in $\bm{y}$ has been contracted yet, the probability that the edge that is selected for contraction in iteration $i$ is in $\bm{y}$ is
            \[
                p_i(e \in \bm{y})
                = \sum_{e \in \bm{y}} p_i(e)
                = \frac{\sum_{e \in \bm{y}} \big( 1 - \sigma(\bm{s}[e]) \big)}{\sum_{e' \in E_i} \big( 1 - \sigma(\bm{s}[e']) \big)}
            \]

            Karger's algorithm outputs a given cut $\bm{y}$ if and only if no edge in $\bm{y}$ is contracted by the algorithm
            (see~\citet{karger-stein}, lemma 2.1).
            Let $k$ be the parameter for minimum $k$-cut, i.e.\ the number of desired connected components.
            Karger's algorithm will always terminate after $|V| - k$ contraction steps.
            The probability that no edge in $\bm{y}$ is contracted during the $|V| - k$ contraction steps is
            \[
                h(\bm{y} \mid \gG, \bm{s})
                = \prod_{i=1}^{|V|-k} \big(1 - p_i(e \in \bm{y}) \big)
                = \prod_{i=1}^{|V|-k} \left(1 - \frac{\sum_{e \in \bm{y}} \big( 1 - \sigma(\bm{s}[e]) \big)}{\sum_{e' \in E_i} \big( 1 - \sigma(\bm{s}[e']) \big)} \right)
            \]

            In the numerator, $\lim_{\sigma(s) \to \bm{y}} \sum_{e \in \bm{y}} \big( 1 - \sigma(\bm{s}[e]) \big) = 0$, since all $\sigma(\bm{s}[e])$ go to 1.
            The denominator is greater than zero, because there is at least one edge left that is not in $\bm{y}$, otherwise the algorithm would be finished.
            Since the fraction goes towards zero, all terms of the product go towards 1 and $\lim_{\sigma(\bm{s}) \to \bm{y}} h(\bm{y} \mid \gG, \bm{s}) = 1$.
        \end{proof}

        The result can trivially be extended to the Karger--Stein algorithm, a variant of Karger's algorithm, and to weighted graphs $\gG = (V, E, w)$.

    \subsection{Probability of finding optimal solutions using algorithms with relative error bounds}

        Let $\gG = (V, E, w)$ be a weighted graph with edge weights $w: E \rightarrow \mathbb{R}_{>0}$.
        Consider any \ac{CO} problem that asks to find a set of edges of minimum total weight that fits the constraints (e.g.\ minimum $k$-cut or \ac{TSP}).
        This means that the \ac{CO} problem's objective function takes the form $J(\hat{\bm{y}}) = \sum_{e \in \hat{\bm{y}}} w(e)$. Remember that we allow ourselves the notational convenience of treating $\hat{\bm{y}}$ as both a set of edges and a vector in $\{0, 1\}^{|E|}$.

        \begin{tcolorbox}
            \begin{theorem}
                Let $J$ be the objective of a minimization problem on graphs with  $J(\hat{\bm{y}}) = \sum_{e \in \hat{\bm{y}}} w(e)$ and $w(e) > 0$. Let $\gG = (V, E, w)$ be an undirected, weighted graph, and let $\bm{y} \in \{0, 1\}^{|E|}$ be an optimal solution on $\gG$.
                Moreover, let $h(\hat{\bm{y}} \mid \gG)$ be a probabilistic approximation algorithm with a bound on the relative error, that is, there exists a function $\alpha$ such that for all $\hat{\bm{y}} \sim h(\hat{\bm{y}} \mid \gG)$, $J(\hat{\bm{y}}) \leq \alpha(\gG) J(\bm{y})$.
                Finally, let $h(\hat{\bm{y}} \mid \gG, \bm{s})$ be the parameterized version of that approximation algorithm, using parameters $\bm{s} \in \mathbb{R}^{|E|}$, and let $\sigma$ be the element-wise sigmoid function.
                Then,
                \[
                    \lim_{\sigma(\bm{s}) \to \bm{y}} h(\bm{y} \mid \gG, \bm{s}) = 1.
                \]
            \end{theorem}
        \end{tcolorbox}

        \begin{proof}
            Let $\gG'$ be the graph that the approximation algorithm receives as input after modification through $\bm{s}$, i.e.\ $h(\hat{\bm{y}} \mid \gG, \bm{s}) = h(\hat{\bm{y}} \mid \gG')$.
            Let $J_{\gG'}(\bm{y})$ denote the total weight of a solution in the modified graph $\gG'$.
            As $\sigma(\bm{s})$ approaches $\bm{y}$, the weights in $\gG'$ of edges in $\bm{y}$ approach 0.
            This means that the total weight of the optimal solution $\bm{y}$ in the modified graph $\gG'$ approaches 0:
            \[
                \lim_{\sigma(\bm{s}) \to \bm{y}} J_{\gG'}(\bm{y}) = 0.
            \]

            Any other solution $\bm{y}'$ cannot be a strict subset of the edges included in $\bm{y}$, since that would mean that it has a lower total weight than $\bm{y}$.
            $\bm{y}'$ therefore includes at least one edge that is not included in $\bm{y}$.
            Since this edge's original edge weight is greater than 0, its scaled weight remains greater than 0 as $\sigma(\bm{s})$ approaches $\bm{y}$.
            The total weight in $\gG'$ of any solution other than $\bm{y}$ therefore approaches a value greater than 0:
            \[
                \lim_{\sigma(\bm{s}) \to \bm{y}} J_{\gG'}(\bm{y}') > 0 \quad \forall \bm{y}' \neq \bm{y}.
            \]

            In the limit, the right hand side of the relative error bound $J_{\gG'}(\hat{\bm{y}}) \leq \alpha(G) J_{\gG'}(\bm{y})$ approaches 0. 
            This means that the maximum total weight of solutions that the approximation algorithm can find approaches 0, and, as soon as the lengths of the suboptimal solutions in $\gG'$ are far enough away from 0, $\bm{y}$ is the only solution left that the approximation algorithm can find.
        \end{proof}

        Note that this proof can trivially be adapted to \ac{CO} problems that asks to find a set of \emph{nodes} of minimum total weight that fits the constraints.

        We know that the \ac{TSP} tour produced by any insertion algorithm $h(\hat{\bm{y}} \mid \gG)$, such as random insertion \citep{random-insertion}, is within $\lceil \log(|V|) \rceil + 1$ of the optimal tour~\citep{insertion-algorithms-approximation-ratio}.
        Formally, this means that $J(\hat{\bm{y}}) \leq \big( \lceil \log(|V|) \rceil + 1 \big) \: J(\bm{y})$ for $\hat{\bm{y}} \sim h(\hat{\bm{y}} \mid \gG)$.
        From that, the following result immediately follows:

        \begin{corollary}
            Let $\gG = (V, E, w)$ be an undirected, weighted graph, and let $\bm{y} \in \{0, 1\}^{|E|}$ be a minimal length TSP tour on $\gG$.
            Let $h(\hat{\bm{y}} \mid \gG, \bm{s})$ be a probabilistic insertion algorithm, guided by parameters $\bm{s} \in \mathbb{R}^{|E|}$.
            Let $\sigma$ be the element-wise sigmoid function.
            Then,
            \[
                \lim_{\sigma(\bm{s}) \to \bm{y}} h(\bm{y} \mid \gG, \bm{s}) = 1.
            \]
        \end{corollary}

\section{Implementation details}
\label{sec:implementation-details}

    \subsection{Residual gated graph convnets}

        We use residual gated graph convnets \citep{residual-gated-graph-convnets}, but adapt them to include edge features $\bm{e}_{ij}^l$ and a dense attention map $\bm{\eta}_{ij}^l$ following \citet{an-efficient-gcn-technique-for-the-tsp}.
        The input node features $\bm{x}_i^0$ and edge features $\bm{e}_{ij}^0$ are first pre-processed using a single-layer \acs{MLP} (\acl{MLP}\acused{MLP}) for each of the two.
        Each further layer is computed as follows:
        \begin{align*}
            \bm{x}_i^{l+1} &= \bm{x}_i^l + \mathrm{ReLU} \Bigg( \mathrm{BN} \bigg( W_1^l \bm{x}_i^l + \sum_{j \in \mathcal{N}_i} \bm{\eta}_{ij}^l \odot W_2^l \bm{x}_j^l \bigg) \Bigg)
            \hspace{2mm} \text{with } \bm{\eta}_{ij}^l = \frac{\sigma \big( \bm{e}_{ij}^l \big)}{\sum_{j' \in \mathcal{N}_i} \sigma \big( \bm{e}_{ij'}^l + \varepsilon \big)} \in \mathbb{R}^d, \\
            \bm{e}_{ij}^{l+1} &= \bm{e}_{ij}^l + \mathrm{ReLU} \Bigg( \mathrm{BN} \bigg( W_3^l \bm{e}_{ij}^l + W_4^l \bm{x}_i^l + W_5^l \bm{x}_j^l \bigg) \Bigg),
        \end{align*}
        where $W_1^l, \dots, W_5^l \in \mathbb{R}^{d \times d}$ are learnable weights, $d$ is the hidden dimension, $\mathrm{ReLU}$ is the rectified linear unit, $\mathrm{BN}$ is batch normalization, $\sigma = \frac{e^x}{1 + e^x}$ is the element-wise sigmoid function, and $\varepsilon$ is an arbitrary small value.
        $\odot$ denotes the Hadamard product, and $\mathcal{N}_i$ denotes the set of nodes that are adjacent to $i$.

        The final edge-level output is calculated from the last layer's edge features $\bm{e}_{ij}^l$ using another \ac{MLP}.
        $f(\gG) \in \mathbb{R}^{|E|}$ refers to applying this \ac{GNN} on a graph $\gG$.

    \subsection{Decoding using the Karger--Stein algorithm}

        In the case of the Karger--Stein algorithm, we noticed empirically that simply modifying the input graph can lead to degenerate behavior during testing.
        The Karger--Stein algorithm uses the graph's edge weights in two places:
        (1) when sampling an edge for contraction and
        (2) when comparing the cuts that resulted from different recursion arms.
        We noticed that the performance of our overall method can be improved when using a model trained with the setting described in \autoref{sec:preference-based-gradient-estimation} by using the modified edge weights for the first case and the original edge weights for the second case.
        Intuitively, if the GNN makes a mistake when scaling the edge weights, using the original edge weights for comparing cuts can allow the Karger--Stein algorithm to find the optimal cut regardless.

\section{Details regarding baselines}
\label{sec:baseline-details}

    \subsection{Supervised training with binary cross entropy loss}

        The task is treated as an edge-level binary classification task.
        The network is trained using a \ac{BCE} loss:
        \begin{align*}
            \hat{y} &= \sigma \big( f_\theta(\gG) \big) \\
            \mathcal{L}_\text{supervised BCE}(\gG, y) &= \text{BCE}(\hat{y}, y) = \sum_{i=1}^{|E|} y_i \log \hat{y}_i + (1 - y_i) \log(1 - \hat{y}_i)
        \end{align*}
        where $f_\theta$ is a \ac{GNN}, $\gG$ is the input graph,
        $y\in \{ 0, 1 \}^{|E|}$ is the ground truth solution and $\hat{y} \in (0, 1)^{|E|}$ is the predicted solution.

        A ground truth label of 1 represents that an edge belongs to a minimum $k$-cut or a \ac{TSP} tour.

    \subsection{REINFORCE}

        Recall that the REINFORCE algorithm \citep{reinforce}, also known as the score function estimator, calculates
        \[
            \nabla_\theta \mathbb{E}_{y \sim p_\theta(x)} \big[ J(y) \big]
            = \mathbb{E}_{y \sim p_\theta(x)} \big[ J(y) \nabla_\theta \log p_\theta(x) \big]
        \]
        where $J$ is an objective function, and $p_\theta(x)$ is a probability distribution parameterized by $\theta$.

        We assume that $p_\theta$ is a discrete constrained exponential family distribution, i.e.
        \[
            p_\theta(x) =
            \begin{cases}
                \displaystyle\frac{\exp \big( \langle x, \theta \rangle \big)}{\sum_{x'} \exp \big( \langle x', \theta \rangle \big)} & \text{if } x \text{ satisfies the constraints} \\
                0 & \text{otherwise}
            \end{cases}
        \]

        For valid $x$,
        \[
            \log p_\theta(x) = \langle x, \theta \rangle - A(\theta),
        \]
        where $A(\theta)$ is the log-partition function
        \[
            A(\theta) = \log \left( \sum_{x' \in \mathcal{C}} \exp \big( \langle x', \theta \rangle \big) \right).
        \]
        Since $\nabla_\theta A(\theta) = \mathbb{E}_{y \sim p_\theta(x)}[y]$, we get
        \[
            \nabla_\theta \log p_\theta(x) = x - \mathbb{E}_{y \sim p_\theta(x)}[y].
        \]
        Inserting this into the REINFORCE formula gives us
        \[
            \nabla_\theta \mathbb{E}_{y \sim p_\theta(x)} \big[ J(y) \big] = \mathbb{E}_{y \sim p_\theta(x)} \bigg[ J(y) \Big( y - \mathbb{E}_{y' \sim p_\theta(x)}[y'] \Big) \bigg].
        \]

        Using the Gumbel-max trick, we sample from $p_\theta$ by sampling $\varepsilon \sim \mathrm{Gumbel}(0, 1)$, then calculating $y := h(\theta + \varepsilon)$.

        Estimating the outer expectation by sampling once and the inner expectation by sampling $N$ times, we arrive at \autoref{alg:reinforce}.

        \begin{algorithm}[h!]
            \caption{REINFORCE}
            \label{alg:reinforce}
            \begin{algorithmic}
                \State \textbf{Input:} distribution parameter $\theta$
                \State $\varepsilon \sim \mathrm{Gumbel}(0, 1)$
                \State $y \gets h(\theta + \varepsilon)$
                \State $\nabla_\theta J(y) \gets J(y) \left( y - \displaystyle\frac{1}{N} \sum_{i=1}^N y_i \right)$ \\
                    \hspace{1.5cm} where $\varepsilon_i \sim \mathrm{Gumbel}(0, 1)$ \\
                    \hspace{2.93cm} $y_i \gets h(\theta + \varepsilon_i)$
                \State \textbf{return} $\nabla_\theta J(y)$
            \end{algorithmic}
        \end{algorithm}

    \subsection{Implicit maximum likelihood estimator (I-MLE)}
    \label{sec:imle-details}

        \acused{I-MLE}
        \Ac{I-MLE} \citep{imle} allows estimating gradients with respect to the parameters of discrete exponential family distributions.
        This can be used to backpropagate through \ac{CO} solvers as follows.
        In the forward pass, perturb the input $\bm{\theta} \in \mathbb{R}^n$ to the \ac{CO} solver using noise $\bm{\epsilon} \sim \rho(\bm{\epsilon})$ sampled from a suitable noise distribution.
        Then run the \ac{CO} solver on the perturbed input $\bm{\theta} + \bm{\epsilon}$, obtaining output $\bm{z}$.

        In the backward pass, assume we know the gradient of the loss w.r.t.\ to $\bm{z}$, $\nabla_{\bm{z}} \mathcal{L}$.
        First, obtain a modified input $\bm{\theta}'$ for which we can expect better outputs compared to $\bm{\theta}$.
        One generally applicable option suggested by \citet{imle} is $\bm{\theta}' = \bm{\theta} - \lambda \nabla_{\bm{z}} \mathcal{L}$, where $\lambda$ is a hyperparameter.
        Using the same noise $\bm{\epsilon}$ as in the forward pass, perturb $\bm{\theta}'$ and run the \ac{CO} solver on $\bm{\theta}' + \bm{\epsilon}$, obtaining $\bm{z}'$.
        Finally, return the estimated gradient $\nabla_{\bm{\theta}} \mathcal{L} \approx \bm{z} - \bm{z}'$.

        This produces biased gradient estimates, but with much smaller variance than REINFORCE.
        To further reduce variance, this procedure can be repeated $S$ times, sampling new noise $\bm{\epsilon}_i \sim \rho(\bm{\epsilon}_i)$ each time and averaging the results.

        We used three noise samples from the Sum-of-Gamma noise distribution suggested by \citet{imle}, using $\kappa = 5$ and $100$ iterations.

        \subsubsection{Supervised training with \acs{I-MLE}}

            The outputs of the \ac{GNN} are used to guide a \ac{CO} approximation algorithm.
            This approximation algorithm outputs a solution to the \ac{CO} problem, which can be compared to the ground truth solution using a Hamming loss.
            During backpropagation, I-MLE~\citep{imle} is used to estimate the gradient of the loss with respect to the \ac{GNN}'s output.
            This setting has no practical benefit over the simple supervised training using a \ac{BCE} loss, but it serves to measure the effectiveness of \ac{I-MLE}.
    
            The training procedure works as follows.
            \begin{align*}
                s &= \sigma \big( f_\theta(\gG) \big) \\
                \hat{y} &= h(\gG, 1 - s) \\
                \mathcal{L}_\text{supervised I-MLE}(\gG, y) &= \frac{1}{|E|} \sum_{i=1}^{|E|} \hat{y}_i (1 - y_i) + (1 - \hat{y}_i) y_i
            \end{align*}
            Note that here, $\hat{y} \in \{ 0, 1 \}^{|E|}$ is guaranteed to be a valid solution to the given \ac{CO} problem.
            \ac{I-MLE} is used to estimate $\frac{\mathrm{d} \mathcal{L}}{\mathrm{d} (1 - s)}$.

        \subsubsection{I-MLE target distribution}

            We're using a custom target distribution for \ac{I-MLE} in the supervised setting.
            This target distribution is similar to the target distribution for \ac{CO} problems presented in~\citep{imle}.
            The idea behind it is to recover the ground truth label from the loss, which is possible when using the Hamming loss:
            \begin{align*}
                \ell(\hat{y}, y)                                   &= \hat{y} (1 - y) + (1 - \hat{y}) y \\
                \frac{\mathrm{d}}{\mathrm{d} \hat{y}} \ell(\hat{y}, y) &= 1 - 2y \\
                y &= \frac{1 - \frac{\mathrm{d}}{\mathrm{d} \hat{y}} \ell(\hat{y}, y)}{2}
            \end{align*}
            The best value for $\theta$ is $1 - y$ (we have to invert it because the input to the approximation algorithm is inverted).%
            \footnote{Using $\theta$ in the same sense as \citet{imle}. In our case, $\theta = 1 - s$}
            With this we arrive at the following target distribution:
            \[
                \theta'
                = 1 - y
                = \frac{1 + \frac{\mathrm{d}}{\mathrm{d} \hat{y}} \mathcal{L}}{2}
            \]

        \subsubsection{Self-supervised training with I-MLE}
            Since we're using a \ac{CO} approximation algorithm that guarantees that its outputs are valid solutions to the \ac{CO} problem, we can use the \ac{CO} problem's objective function as a loss directly instead of the supervised Hamming loss.
            The ground truth labels are therefore no longer required.
    
            In the case of minimum $k$-cut, the size of the cut is used as loss function.
            For TSP, the length of the tour is used.
    
            \begin{align*}
                s &= \sigma \big( f_\theta(\gG) \big) \\
                \hat{y} &= h(\gG, 1 - s) \\
                \mathcal{L}_\text{self-supervised I-MLE}(\gG) &= J_\text{CO}(\hat{y}) \\
            \end{align*}
            where $J_\text{CO}$ is the objective function of the \ac{CO} problem.
            Note that the \ac{CO} problem's constraints do not explicitly appear here, because the \ac{CO} approximation algorithm already guarantees that the constraints are met.
    
            As before, \ac{I-MLE} is used to estimate $\frac{\mathrm{d} \mathcal{L}}{\mathrm{d} (s - 1)}$.
            In this setting, the general-purpose target distribution for \ac{I-MLE} presented in~\citep{imle} is used,
            setting $\lambda = 20$ as suggested by that paper.

\section{Graph generation}
\label{sec:graph-generation-appendix}

    \subsection{Minimum \texorpdfstring{$k$}{k}-cut}
    \label{sec:graph-generation-appendix-minimum-k-cut}

        Many commonly used graph generators create graphs with low-degree nodes.
        These graphs contain trivial solutions to the minimum $k$-cut problem in which $k - 1$ connected components only contain one node, and one connected component contains all of the remaining nodes.
        When creating a dataset for minimum $k$-cut, care must therefore be taken to avoid graphs with low-degree nodes.

        \paragraph{Graphs without edge weights.}
        A simple method to generate graphs with meaningful solutions to the minimum $k$-cut problem is as follows.
        Create $k$ fully connected subgraphs of random sizes within a given range.
        Then, add a random number of edges between random nodes of different subgraphs while ensuring that the resulting graph is connected.
        An additional benefit of this method is that, if the number of edges added between subgraphs is smaller than the number of nodes in the smallest subgraph by at least two, then the minimum $k$-cut is known from the construction:
        the cut consists of exactly the edges that were added between subgraphs.
        However, since all graphs generated this way consist of fully connected subgraphs, these problem instances are limited in diversity.

        The range of possible problem instances can be improved by generating graphs of varying density.
        Start by assigning nodes to $k$ subgraphs of random sizes within a given range.
        Then, add a random number of edges that connect nodes of different subgraphs.
        For each subgraph, add edges between random nodes within the same subgraph until all nodes have a higher degree than the number of edges between subgraphs.
        As long as there are enough edges between subgraphs, the minimum $k$-cut very likely consists of the edges between subgraphs.
        The minimum node degree and hence the density of the graph depends on the number of edges between subgraphs and therefore on the size of the minimum $k$-cut.

        \paragraph{Graphs with edge weights.}
        For minimum $k$-cut graphs with edge weights, a graph generator commonly called NOIgen~\citep{noigen} (named after the initials of the authors) is often used.
        NOIgen works by first creating a specified number of nodes and adding edges between random nodes until a specified density is reached (sometimes, a Hamilton path is created first to ensure that the graph is connected).
        The weights of the edges are chosen uniformly at random.
        Finally, the nodes are randomly divided into $k$ subgraphs.
        The weights of edges that connect nodes of different subgraphs are scaled down by a fixed factor.

        When testing traditional, non-learned algorithms, the scaling factor is sometimes chosen to be very small, such that the minimum $k$-cut is very likely to consist of the edges between subgraphs~\citep{experimental-study-of-minimum-cut-algorithms}.
        However, this makes the problem trivially easy for GNNs, which can learn that a very low edge weight corresponds to an edge belonging to the minimum $k$-cut.
        This allows the GNN to disregard the graph structure and therefore circumvent the challenging part of the problem.
        On the other hand, if the weights of edges between subgraphs are not scaled down enough, the generated graph might have a trivial solution that simply cuts out $k - 1$ nodes.

        To combat this problem, we modify NOIgen by controlling not just the weights of edges between subgraphs, but also the number of edges between subgraphs.
        We add a parameter that specifies which fraction of edges is generated between subgraphs (as opposed to within the same subgraph).
        Ensuring that there are few enough edges between subgraphs allows for a milder downscaling of their edge weights without introducing a trivial solution.
        This in turn prevents the GNN from inferring whether an edge belongs to the minimum $k$-cut simply from its weight.

        The minimum $k$-cut in these graphs usually consists of the edges between subgraphs, but this is not guaranteed.
        The ground truth solution is therefore calculated separately to make sure that it reflects the optimal cut, as described in \autoref{sec:calculating-ground-truth-labels-appendix}.
        Another benefit of calculating them separately is that we can set the number of subgraphs to a different number than $k$, generating more interesting graphs.

    \subsection{Travelling salesman problem}

        Instances of Euclidean \ac{TSP} are commonly generated with this simple algorithm:
        \begin{enumerate}
            \item Create a fully connected graph with $n$ nodes
            \item For each node, draw a position in the unit square uniformly at random and assign it as node features
            \item Calculate the distances between the nodes and assign them as edge features
        \end{enumerate}

\newpage
    \subsection{Calculating ground truth labels for supervised training}
    \label{sec:calculating-ground-truth-labels-appendix}

        In general, ground truth labels are generated using a traditional (i.e.\ non-learned) algorithm.
        In some settings, the graph can be constructed such that the ground truth solution can be obtained simultaneously from the same construction process, in which case running the traditional algorithm is not necessary.
        \begin{itemize}
            \item Minimum k-cut: For graphs without edge weights, the graphs can be constructed with known ground truth solutions.
            See \autoref{sec:graph-generation-appendix-minimum-k-cut} for details.
            For graphs with edge weights, the Karger--Stein algorithm~\citep{karger-stein} is run 100 times,
            and the smallest cut found is treated as the ground truth minimum cut.
            \item TSP: The well-established Concorde solver~\citep{concorde}, which guarantees optimal solutions, is used to generate ground truth labels.
        \end{itemize}

    \begin{table*}
    \centering
    \caption{
        TSP optimality gaps with mean $\pm$ standard deviation calculated over ten evaluation runs on the same model parameters.
        We group methods by their original paper and indicate the decoder used in parentheses.
        Results marked with * are values obtained from the indicated papers and therefore do not include standard deviations.
    }
    \label{tab:tsp-optimality-gaps-extended}
    \small
    \resizebox{\textwidth}{!}{%
        \begin{tabular}{llll}
            \toprule
            \textbf{Method} (decoder in parentheses)   & $\bm{n = 20}$   & $\bm{n = 50}$   & $\bm{n = 100}$  \\
            \midrule
            \textbf{Self-supervised}                   &                 &                 &                 \\
            \citet{neural-co-with-rl}                  &                 &                 &                 \\
            \quad (greedy)                             & 1.42\%*         & 4.46\%*         & 6.90\%*         \\
            \citet{learning-heuristics-for-the-tsp}    &               &                 &                 \\
            \quad (greedy)                             & 0.66\%* (2m)    & 3.98\%* (5m)    & 8.41\%* (8m)    \\
            \quad (greedy + 2OPT)                      & 0.42\%* (4m)    & 2.77\%* (26m)   & 5.21\%* (3h)    \\
            \quad (sampling 1280 times)                & 0.11\%* (5m)    & 1.28\%* (17m)   & 12.70\%* (56m)  \\
            \quad (sampling 1280 times + 2OPT)         & 0.09\%* (6m)    & 1.00\%* (32m)   & 4.64\%* (5h)    \\
            \citet{learning-co-algorithms-over-graphs} &                 &                 &                 \\
            \quad (greedy)                             & 1.42\%*         & 5.16\%*         & 7.03\%*         \\
            \citet{attention-learn-to-solve-routing-problems} &          &                 &                 \\
            \quad (greedy)                             & 0.34\%*         & 1.76\%* (2s)    & 4.53\%* (6s)    \\
            \quad (sampling 1280 times)                & 0.08\%* (5m)    & 0.52\%* (24m)   & 2.26\%* (1h)    \\
            REINFORCE (random ins., 20 runs)       & 7.98\%$\pm$0.08 (769ms)       & 23.84\%$\pm$0.10 (11.62s) & 52.83\%$\pm$0.47 (1.38m) \\
            REINFORCE (random ins., 100 runs)      & 4.78\%$\pm$0.03 (3.64s)       & 11.24\%$\pm$0.08 (57.85s) & 31.06\%$\pm$0.25 (7.53m) \\
            I-MLE (random ins., 20 runs)           & 10.54\%$\pm$0.15 (761ms)      & 35.87\%$\pm$0.34 (11.14s) & 61.42\%$\pm$0.23 (1.37m) \\
            I-MLE (random ins., 100 runs)          & 6.55\%$\pm$0.10 (3.68s)       & 19.40\%$\pm$0.24 (56.63s) & 44.42\%$\pm$0.14 (7.51m) \\
            Best-of-20 (random insertion, 20 runs)      & 0.40\%$\pm$0.03 (787ms)          & 11.94\%$\pm$0.11 (11.03s) & 15.82\%$\pm$0.09 (1.47m) \\
            Best-of-20 (random insertion, 100 runs)     & 0.10\%$\pm$0.01 (3.70s)          & 5.21\%$\pm$0.15 (52.82s)  & 9.82\%$\pm$0.09 (7.38m)  \\
            PBGE \textbf{(ours)} (random ins., 20 runs)  & 0.18\%$\pm$0.01 (763ms) & 2.37\%$\pm$0.02 (11.11s)  & 5.12\%$\pm$0.06 (1.43m) \\
            PBGE \textbf{(ours)} (random ins., 100 runs) & 0.05\%$\pm$0.01 (3.73s) & 1.13\%$\pm$0.03 (53.98s)  & 3.67\%$\pm$0.05 (7.44m) \\
            \midrule
            \textbf{Supervised}, not directly comparable &                 &                 &                 \\
            \citet{an-efficient-gcn-technique-for-the-tsp} &             &                 &                 \\
            \quad (greedy)                             & 0.60\%* (6s)    & 3.10\%* (55s)   & 8.38\%* (6m)    \\
            \quad (beam search, beam width 1280)            & 0.10\%* (20s)   & 0.26\%* (2m)    & 2.11\%* (10m)   \\
            \quad (beam search, width 1280 + heuristic)       & 0.01\%* (12m)   & 0.01\%* (18m)   & 1.39\%* (40m)   \\
            \citet{difusco}                            &                 &                 &                 \\
            \quad (greedy)                             &                 & 0.10\%*         & 0.24\%*         \\
            \quad (sampling 16 times)                  &                 & 0.00\%*         & 0.00\%*         \\
            \ac{BCE} loss (random insertion, 20 runs)  & 0.15\%$\pm$0.01 (787ms)          & 0.95\%$\pm$0.03 (10.92s) & 2.86\%$\pm$0.04 (1.47m) \\
            \ac{BCE} loss (random insertion, 100 runs) & 0.04\%$\pm$0.00 (3.75s)          & 0.59\%$\pm$0.02 (54.79s) & 1.75\%$\pm$0.03 (7.35m) \\
            \midrule
            \multicolumn{2}{l}{\textbf{Non-learned approximation algorithms}} &                 &                 \\
            Christofides                               & 8.72\%$\pm$0.00 (45ms) & 11.07\%$\pm$0.00 (685ms) & 11.86\%$\pm$0.00 (4.45s) \\
            Random Insertion                           & 4.46\%$\pm$0.08 (41ms) & 7.57\%$\pm$0.08 (575ms)  & 9.63\%$\pm$0.11 (4.34s) \\
            Farthest Insertion                         & 2.38\%$\pm$0.00 (57ms) & 5.50\%$\pm$0.00 (909ms)  & 7.58\%$\pm$0.00 (5.68s) \\
            LKH3                                       & 0.00\%* (18s)          & 0.00\%* (5m)             & 0.00\%* (21m)   \\
            \midrule
            \textbf{Exact solvers}, not directly comparable &                 &                 &                 \\
            Concorde~\citep{concorde}                  & 0.00\%* (1m)    & 0.00\%* (2m)    & 0.00\%* (3m)    \\
            Gurobi                                     & 0.00\%* (7s)    & 0.00\%* (2m)    & 0.00\%* (17m)   \\
            \bottomrule
    \end{tabular}}
\end{table*}

\newpage
\section{Extended comparison for TSP}
\label{sec:extended-comparison}

    \autoref{tab:tsp-optimality-gaps-extended} compares \ac{PBGE} with more baselines, including ones that make use of supervised learning.

    We also compare against a simple self-supervised baseline that runs random insertion on the input graph 20 times and treats the best solution found as ground truth for a \ac{BCE} loss.
    We call this baseline ``Best-of-20''.

    The beam search decoder works similarly to the greedy decoder.
    It starts with an arbitrary node, then explores the $b$ edges with the highest scores.
    This gives us $b$ partial solutions.
    In each iteration, each partial solution is expanded at its last node, and out of the resulting paths, the $b$ best partial solutions are kept.
    Edges that would lead to invalid tours are ignored.
    The parameter $b$ is called the \emph{beam width}, and beam search with $b = 1$ corresponds to greedy search.

\section{Hyperparameters and other details}
\label{sec:hyperparameters}

    \autoref{tab:hyperparameters-min-k-cut} and \autoref{tab:hyperparameters-tsp} detail the hyperparameters used for experiments on minimum $k$-cut and \ac{TSP}, respectively.

    Our training sets for minimum $k$-cut contain 10,000 graphs, and the validation sets contain 1,000 graphs.
    For \ac{TSP}, our training sets consist of 1,000,000 graphs, and our validation sets contain 1,000 graphs.

    \begin{table*}
        \centering
        \caption{Hyperparameters used for minimum $k$-cut.}
        \label{tab:hyperparameters-min-k-cut}
        \begin{tabular}{ll}
            \toprule
            Hyperparameter                   & Value             \\
            \midrule
            \ac{GNN} layers                  & 4                 \\
            \ac{GNN} hidden channels         & 32                \\
            \ac{MLP} prediction head layers  & 2                 \\
            \midrule
            Optimizer                        & AdamW             \\
            Weight Decay                     & 0.01              \\
            Learning rate scheduler          & ReduceLROnPlateau \\
            Initial learning rate            & 0.001             \\
            Learning rate scheduler patience & 4                 \\
            Batch size                       & 64 graphs         \\
            \bottomrule
        \end{tabular}
    \end{table*}

    \begin{table*}
        \centering
        \caption{Hyperparameters used for \ac{TSP}.}
        \label{tab:hyperparameters-tsp}
        \begin{tabular}{ll}
            \toprule
            Hyperparameter                   & Value             \\
            \midrule
            \ac{GNN} layers                  & 12                \\
            \ac{GNN} hidden channels         & 100               \\
            \ac{MLP} prediction head layers  & 3                 \\
            \midrule
            Optimizer                        & AdamW             \\
            Weight Decay                     & 0.01              \\
            Learning rate scheduler          & ReduceLROnPlateau \\
            Initial learning rate            & 0.0001            \\
            Learning rate scheduler patience & 4                 \\
            Batch size                       & 64 graphs         \\
            \bottomrule
        \end{tabular}
    \end{table*}

\section{Hardware resources used for experiments}
\label{sec:hardware-resources}

    We used an AMD EPYC 7313 16-Core processor and an NVIDIA H100 GPU in our experiments.
    Runtime measurements were taken on an 11th Gen Intel Core i7-11850H processor with a base speed of 2.50\,GHz, using a single CPU core and no GPU, with a batch size of 1.

\end{document}